\documentclass[conference]{IEEEtran}
\IEEEoverridecommandlockouts
\usepackage{cite}
\usepackage{amsmath,amssymb,amsfonts}
\usepackage{algorithmic}
\usepackage{graphicx}
\usepackage{textcomp}
\usepackage{xcolor}
\usepackage{enumerate}
 \usepackage{hyperref}
\usepackage{mathrsfs}
\usepackage{bm}
\usepackage{url}
\usepackage{xspace}
\usepackage{subcaption}
\usepackage{tabularray}

\def\BibTeX{{\rm B\kern-.05em{\sc i\kern-.025em b}\kern-.08em
    T\kern-.1667em\lower.7ex\hbox{E}\kern-.125emX}}
\begin{document}

\title{CLAPS: A CLIP-Unified Auto-Prompt Segmentation for Multi-Modal Retinal Imaging\\
% {\footnotesize \textsuperscript{*}Note: Sub-titles are not captured for https://ieeexplore.ieee.org  and
}

% \author{\IEEEauthorblockN{1\textsuperscript{st} Anonymized Authors}
% \IEEEauthorblockA{\textit{Anonymized Affiliations)} \\
% \textit{Anonymized Organization}\\
% email@anonymized.com}
% \and
% \and
% \and
% \and
% \and

\author{\IEEEauthorblockN{Zhihao Zhao}
\IEEEauthorblockA{\textit{Technical University of Munich }\\
Munich, Germany \\
zhihao.zhao@tum.de}
\and
\IEEEauthorblockN{Yinzheng Zhao}
\IEEEauthorblockA{\textit{ Technical University of Munich }\\
Munich, Germany \\
yinzheng.zhao@tum.de}
\and
\IEEEauthorblockN{Junjie Yang}
\IEEEauthorblockA{\textit{ Technical University of Munich }\\
Munich, Germany \\
junjie.yang@tum.de}
\and
\IEEEauthorblockN{Xiangtong Yao}
\IEEEauthorblockA{\textit{ Technical University of Munich }\\
Munich, Germany  \\
xiangtong.yao@tum.de}
\and
\IEEEauthorblockN{Quanmin Liang}
\IEEEauthorblockA{\textit{ Sun Yat-Sen University }\\
Guangzhou, China  \\
liangqm5@mail2.sysu.edu.cn}
\and
\IEEEauthorblockN{Shahrooz Faghihroohi}
\IEEEauthorblockA{\textit{ Technical University of Munich }\\
Munich, Germany  \\
shahrooz.faghihroohi@tum.de}
\and
\IEEEauthorblockN{Kai Huang}
\IEEEauthorblockA{\textit{ Sun Yat-Sen University }\\
Guangzhou, China \\
huangk36@mail.sysu.edu.cn}
\and
\IEEEauthorblockN{Nassir Navab}
\IEEEauthorblockA{\textit {Technical University of Munich }\\
Munich, Germany \\
nassir.navab@tum.de}
\and
\IEEEauthorblockN{M.Ali Nasseri}
\IEEEauthorblockA{\textit{Technical University of Munich}\\
Munich, Germany \\
ali.nasseri@mri.tum.de}
}

\maketitle
\newcommand{\CLAPS}{{\textbf{CLAPS}}\xspace}
\begin{abstract}
Recent advancements in foundation models, such as the Segment Anything Model (SAM), have significantly impacted medical image segmentation, especially in retinal imaging, where precise segmentation is vital for diagnosis. Despite this progress, current methods face critical challenges: 1) modality ambiguity in textual disease descriptions, 2) a continued reliance on manual prompting for SAM-based workflows, and 3) a lack of a unified framework, with most methods being modality- and task-specific.
To overcome these hurdles, we propose \textbf{CL}IP-unified \textbf{A}uto-\textbf{P}rompt \textbf{S}egmentation (\CLAPS), a novel method for unified segmentation across diverse tasks and modalities in retinal imaging. Our approach begins by pre-training a CLIP-based image encoder on a large, multi-modal retinal dataset to handle data scarcity and distribution imbalance. We then leverage GroundingDINO to automatically generate spatial bounding box prompts by detecting local lesions. To unify tasks and resolve ambiguity, we use text prompts enhanced with a unique "modality signature" for each imaging modality. Ultimately, these automated textual and spatial prompts guide SAM to execute precise segmentation, creating a fully automated and unified pipeline. Extensive experiments on 12 diverse datasets across 11 critical segmentation categories show that CLAPS achieves performance on par with specialized expert models while surpassing existing benchmarks across most metrics, demonstrating its broad generalizability as a foundation model.
\end{abstract}


\section{Introduction}
The integration of artificial intelligence in medical image analysis has led to remarkable progress in various tasks, including segmentation, classification, and detection \cite{kulyabin2024segment,li2025adapting}. In ophthalmology, accurate segmentation of anatomical structures and pathological regions is critical for diagnosis and treatment planning. With the advent of multi-modal imaging techniques, clinicians now have access to complementary information from various sources \cite{mansoor2014generic}. However, this diversity also poses significant challenges—most notably, the scarcity of annotated data and the imbalance across different imaging modalities. These issues complicate their tendency for performance degradation when handling inherently imbalanced multi-modal data (as evidenced by the typical 10:1 ratio of OCT to ICGA samples \cite{feucht2016oct}) and their fundamental failure to effectively leverage cross-modal semantic relationships that are crucial for comprehensive diagnostic analysis. Additionally, a key challenge in multi-modal imaging is modality ambiguity in textual disease descriptions. Specifically, many ophthalmic diseases are described by universal texts, yet they exhibit starkly different visual features across various imaging modalities.

Current solutions are integrated into foundation models to learn a unified paradigm on large-scale datasets.
In small datasets, these methods exacerbate data imbalance issues and increase deployment complexity. While foundation models like SAM offer promising zero-shot segmentation capabilities, despite their impressive capabilities, the direct application of these models to medical imaging suffers from critical limitations \cite{sun2024medical,khan2025comprehensive}, particularly in their reliance on expert-curated prompts, which significantly hinders clinical workflow automation \cite{gan2025review}. MedCLIP-SAM \cite{koleilat2024medclip} is capable of addressing the challenge of fully automated segmentation in SAM. Nevertheless, CAM-based approaches have inherent limitations for segmentation. They fail to provide the pixel-level accuracy required for fine-grained tasks and typically struggle to distinguish and segment multiple, distinct instances of lesions present in the same image.

Our methodology is designed to create a versatile and robust framework for retinal lesion segmentation, addressing several key challenges in automated medical image analysis.
First, to overcome the dependency of the Segment Anything Model (SAM) \cite{ma2024segment} on manual interaction, we automate the prompting process. We employ GroundingDINO \cite{liu2024grounding} to automatically localize regions of interest (ROIs) within the images, generating precise spatial bounding box prompts. These prompts then guide SAM to perform accurate and context-aware delineation of retinal lesions.
Furthermore, to enhance the model's adaptability across diverse imaging modalities, datasets, and segmentation tasks, we enable it to be guided by textual descriptions. This allows for the segmentation of all relevant lesions throughout an entire image based on a single text command. However, this introduces the challenge of Modality Ambiguity: a single clinical term may correspond to vastly different visual features across modalities. For example, "retinal thickening" manifests as an increased thickness of retinal layers in Optical Coherence Tomography (OCT) images but may appear as a subtle, ill-defined discoloration with no depth cues in color fundus photographs or even be imperceptible. To resolve this, we introduce a novel Feature Fusion Module with a Modality Signature. This module preemptively injects modality-specific context into the text feature vector before its fusion with the image features, thereby disambiguating the textual prompt.
Finally, a cornerstone of our approach is addressing the persistent challenges of limited data availability and imbalanced modality distributions. To this end, we pre-train a robust image encoder using CLIP \cite{conde2021clip}, capitalizing on its exceptional ability to map images and text into a unified feature space. By aligning multi-modal retinal images in this shared space, our framework not only enhances segmentation accuracy but also significantly improves its robustness to variations in image quality and modality-specific characteristics.

The major contributions of this work are summarized as follows: Firstly, we developed an automated retinal segmentation pipeline that integrates GroundingDINO with the Segment Anything Model (SAM), replacing manual interaction with automatically generated bounding box prompts for efficient and scalable lesion segmentation.
Secondly, a novel feature fusion module with a modality signature was proposed to resolve 'Modality Ambiguity' by injecting modality-specific context into textual descriptions, enabling accurate lesion segmentation across diverse imaging types from a single text prompt.
Finally, we leverage CLIP-based pre-training to address data scarcity by creating a unified feature space that aligns multi-modal retinal images, significantly improving model robustness and generalization in low-data settings.

\section{Related Work}

\subsection{Task-Specific Segmentation in Ophthalmology}
The advent of deep learning, particularly Convolutional Neural Networks (CNNs), has marked a paradigm shift. Architectures based on the U-Net \cite{ronneberger2015u} and its numerous variants (e.g., UNet++ \cite{zhou2018unet++}, Attention U-Net \cite{oktay2018attention}) have become the  standard for medical image segmentation. These models have demonstrated state-of-the-art performance in segmenting retinal vessels from fundus photographs, delineating fluid regions in OCT scans, and identifying microaneurysms \cite{tahir2024advances}. However, these methods are designed for task-specific segmentation within a single modality, which requires researchers to train or fine-tune models—a process that is highly time-consuming \cite{butoi2023universeg}. Therefore, achieving unified medical image segmentation remains a significant challenge \cite{tang2022unified}.

\subsection{Cross-Modal Learning with Vision-Language Pre-training in Ophthalmology}

The application of Vision-Language Pre-training (VLP) in ophthalmology is a nascent but rapidly growing field. Initial studies have explored pairing ophthalmic images with their associated clinical text to move beyond simple classification. Researchers have demonstrated the feasibility of zero-shot disease classification on retinal fundus images by adapting the CLIP framework to the ophthalmic domain \cite{wang2025osam,baliah2023exploring}. Beyond classification, other cross-modal applications are emerging, such as Visual Question Answering (VQA) \cite{chen2023chatffa}, where models answer natural language questions about an image's content. A significant step towards a foundation model for ophthalmology was the development of RETFound \cite{zhou2023foundation}, a self-supervised model pre-trained on 1.6 million retinal images, which provides a powerful visual backbone for various downstream tasks.

\subsection{SAM in Ophthalmic Image Analysis}
Recently, the Segment Anything Model (SAM) \cite{kirillov2023segment} has shown a powerful capability for zero-shot segmentation of objects in natural images. Inspired by the success of these models, researchers have begun to explore the application of SAM to medical image analysis.
However, as several studies \cite{ali2025review,fan2025research} have pointed out, the direct application of SAM to the medical domain faces key limitations. 
To bridge the gap between foundation models and clinical automation, recent research has focused on automating the prompt generation process. A promising direction is to leverage language as a natural interface for segmentation. MedCLIP-SAM \cite{koleilat2024medclip} is a notable attempt in this area, which uses a vision-language model (MedCLIP) to generate text-based prompts for SAM. It transforms text descriptions into coarse localization maps via Class Activation Mapping (CAM). However, this CAM-based localization method is particularly ill-suited for the small, scattered, and irregularly shaped pathological lesions found in ophthalmology, such as microaneurysms or small fluid accumulations.

\section{Methodology}
\begin{figure*}[ht]
	\centering
	\includegraphics[width=0.95\textwidth]{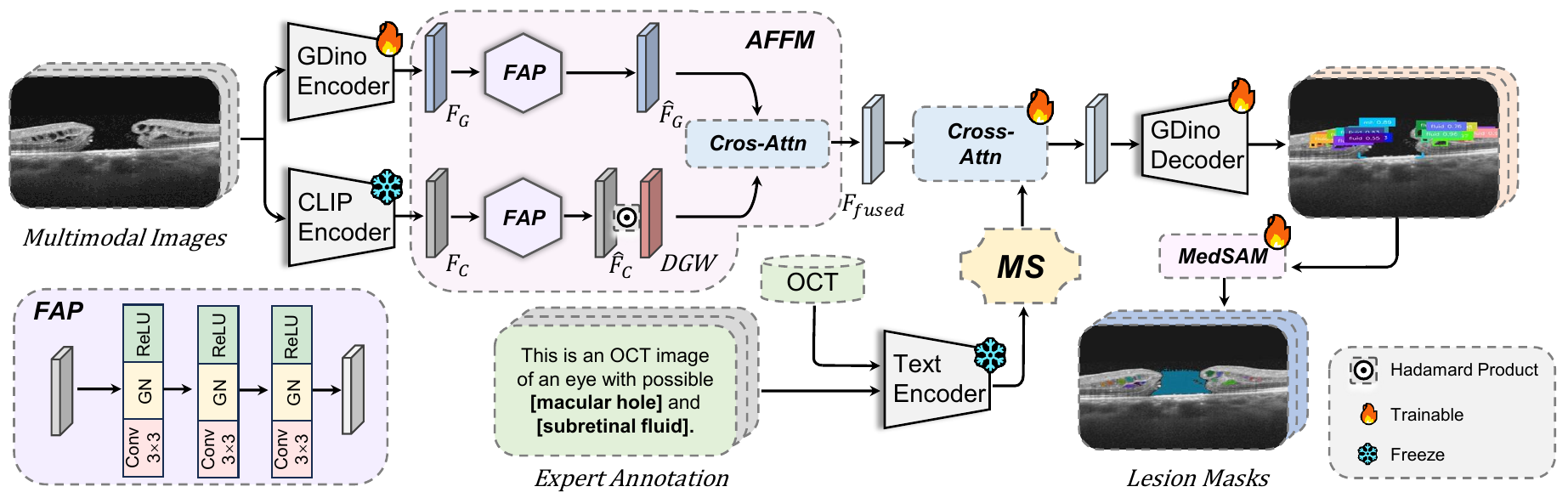}
	\caption{Overview of our proposed \CLAPS. Multi-modal retinal images are initially fused with feature information extracted by the CLIP encoder. Next, an Adaptive Feature Fusion Module (AFFM) fuses the global features from CLIP with the local features from GDino. Simultaneously, text and modality information are integrated through a Modality Signature (MS) module. Then, bounding box prompts are detected using the GroundingDINO module, and finally, fully automated unified segmentation is performed via MedSAM.} 
	\label{fig:flowchart}
\end{figure*}

Fig.~\ref{fig:flowchart} illustrates the comprehensive framework of our proposed model. The architecture begins with the pretraining of a CLIP model \cite{conde2021clip} on an extensive dataset. Subsequently, both the image and text encoders of CLIP are frozen to maintain their learned representations. During the training phase, multimodal retinal images are simultaneously processed through the image encoders of both CLIP and GroundingDino \cite{liu2024grounding}, generating latent features $F_{G}$ and, $F_{C}$ respectively. In the Adaptive Feature Fusion Module (AFFM), these features are then transformed into a unified latent space of identical dimension through the Feature Alignment and Projection (FAP) module \cite{huang2021fapn}, resulting in $\hat{F_{G}}$ and $\hat{F_{C}}$. Notably, $\hat{F_{C}}$ undergoes an element-wise product operation with a learnable matrix of corresponding dimensions, which we designate as the Dynamic Gating Weight (DGW). This parameter matrix essentially learns the dynamic weighting between $\hat{F_{G}}$ and$\hat{F_{C}}$. The subsequent feature fusion is accomplished through CrossAttention, yielding the fused representation $F_{fused}$. To handle the problem of unified training for different modalities, we use a Modality Signature (MS) module to fuse modality information and text prompt information. The architecture then incorporates Grounding's FeatureEnhance module and Decoder module to generate lesion detection bounding boxes. These detected boxes serve as prompts that are sequentially fed into the MedSAM \cite{ma2024segment} for subsequent segmentation tasks.

\subsection{Pretrained CLIP}
We initially pre-trained CLIP on a large-scale and modality-balanced dataset. Specifically, we integrated  publicly available datasets along with internal private fundus images from four hospitals in China and Germany. After removing lower-quality images, resulting in a final collection of 50000 high-quality fundus images and 50000 OCT B-scans.
Given that the available expert annotations in current ophthalmology datasets are predominantly in the form of classification labels rather than natural language text, we enhanced the expert annotations in our dataset by using LLMs to generate 10 candidate templates, and then we randomly selected one during the training phase.
Subsequently, we utilized UniMed-CLIP \cite{uzair2024unimed}  to perform self-supervised training on all fundus and OCT images with expert annotations by contrastive learning strategy.

\subsection{Bounding Box Detection}
In the bounding box detection phase, our objective is to identify and localize specific lesion areas within images based on given textual descriptions. To achieve this, we employ GroundingDino \cite{liu2024grounding}, an advanced object detection model built on the transformer architecture. Since GroundingDino is trained on large-scale datasets, it is susceptible to catastrophic forgetting when fine-tuned on extremely limited datasets. To address this, we integrate UniMed-CLIP's pre-trained encoder during feature extraction. The primary reason for this integration is to leverage CLIP's ability to extract robust, cross-modal general visual features. Trained via contrastive learning on a massive scale, CLIP excels at mapping images and text into a shared, well-aligned feature space. This provides a strong, generalized feature foundation that prevents the model from losing its prior knowledge during fine-tuning on limited data. Furthermore, CLIP's contrastive loss function ensures that features from different modalities are uniformly distributed, which is crucial for reducing feature shift and enhancing the model's performance in complex multimodal tasks.

\subsection{Adaptive Feature Fusion Module (AFFM)}
To effectively integrate the rich semantic information from the CLIP visual encoder with the location-aware features from the GroundingDINO backbone, we introduce an \textbf{Adaptive Feature Fusion Module}. This module is designed to reconcile the inherent heterogeneity between the two feature spaces, CLIP's features being optimized for general visual-language alignment and GroundingDINO's for object detection. Our module achieves this through a two-stage process: (1) a \textbf{Feature Alignment and Projection (FAP)} block that standardizes the feature dimensions, and (2) a \textbf{Gated Cross-Attention Fusion} mechanism that adaptively weights and merges the features.

\subsubsection*{Feature Alignment and Projection (FAP)}

The first step is to preprocess the features from both encoders to ensure they are compatible for fusion. Let $F_{\text{C}} \in \mathbb{R}^{H_c \times W_c \times d_c}$ represent the feature map from the CLIP encoder and $F_{\text{G}} \in \mathbb{R}^{H_g \times W_g \times d_g}$ be the feature map from the GroundingDINO encoder, where $H, W, d$ denote the height, width, and channel dimensions, respectively. These raw features possess different channel depths ($d_c \neq d_g$) and may have disparate spatial resolutions.

The FAP module processes each feature map to extract richer local context and project them into a common, dimensionally consistent space. This is achieved through a sequence of convolutional layers.

\textit{Step 1: Spatial Feature Enhancement}
We first apply two sequential 3x3 convolutional layers to each feature map. These layers, complete with activation functions and normalization (not shown for brevity), serve to expand the receptive field and refine the spatial features within each modality.
\begin{equation}
F'_{\text{C}} = \text{Conv}_{3 \times 3}(\text{ReLU}(\text{Conv}_{3 \times 3}(F_{\text{C}})))
\end{equation}
\begin{equation}
F'_{\text{G}} = \text{Conv}_{3 \times 3}(\text{ReLU}(\text{Conv}_{3 \times 3}(F_{\text{G}})))
\end{equation}

\textit{Step 2: Dimensionality Projection}
Next, a 1x1 convolution is applied to each enhanced feature map. This operation performs a channel-wise linear transformation, projecting both $F'_{\text{C}}$ and $F'_{\text{G}}$ into a shared target dimension, $d_{\text{model}}$. This ensures that the features are dimensionally aligned for the subsequent fusion step.
\begin{equation}
\hat{F}_{\text{C}} = \text{Conv}_{1 \times 1}(F'_{\text{C}}) \quad \in \mathbb{R}^{H_c' \times W_c' \times d_{\text{model}}}
\end{equation}
\begin{equation}
\hat{F}_{\text{G}} = \text{Conv}_{1 \times 1}(F'_{\text{G}}) \quad \in \mathbb{R}^{H_g' \times W_g' \times d_{\text{model}}}
\end{equation}
The resulting features, $\hat{F}_{\text{C}}$ and $\hat{F}_{\text{G}}$, now share the same channel depth and are ready for fusion.

\subsubsection*{Gated Cross-Attention Fusion}
A simple summation or concatenation of the aligned features would treat all information equally. However, the contribution of CLIP's general semantic features should be context-dependent. To this end, we introduce a gating mechanism coupled with a cross-attention block to achieve a more sophisticated, adaptive fusion.

\textit{Step 1: Adaptive Channel Gating}
We introduce a learnable gating mask, $W_{\text{gate}} \in \mathbb{R}^{1 \times 1 \times d_{\text{model}}}$, which is applied to the aligned CLIP features $\hat{F}_{\text{C}}$. This mask adaptively re-weights the feature channels, allowing the model to learn to selectively emphasize or suppress certain semantic concepts from CLIP during the fusion process. The gating operation is performed via element-wise (broadcasted) multiplication:
\begin{equation}
\hat{F}_{\text{C, gated}} = \sigma(W_{\text{gate}}) \odot \hat{F}_{\text{C}}
\end{equation}
Here, $\sigma$ represents the sigmoid function, which normalizes the weights to a range of (0, 1), and $\odot$ denotes element-wise multiplication.

\textit{Step 2: Cross-Attention Fusion}
The final fusion is performed using a standard cross-attention mechanism. This allows the GroundingDINO features to "query" the gated CLIP features, integrating the most relevant semantic context. The GroundingDINO features provide the Query (Q), while the gated CLIP features provide the Key (K) and Value (V).

First, the Query, Key, and Value vectors are linearly projected from their respective feature maps:
\begin{equation}
 \begin{aligned}
Q &= \text{W}_Q \hat{F}_{\text{G}} \\
K &= \text{W}_K \hat{F}_{\text{C, gated}} \\
V &= \text{W}_V \hat{F}_{\text{C, gated}}
\end{aligned}   
\end{equation}

The attention scores are then computed, followed by the weighted summation of the Value vectors:
\begin{equation}
\text{Cross-Attn}(Q, K, V) = \text{softmax}\left(\frac{QK^T}{\sqrt{d_k}}\right)V
\end{equation}
where $d_k$ is the dimension of the key vectors. The output of the attention block is then added back to the original GroundingDINO features via a residual connection, allowing the fusion to act as an enhancement:
\begin{equation}
F_{\text{fused}} = \hat{F}_{\text{G}} + \text{Cross-Attn}(Q, K, V)
\end{equation}
The resulting feature map, $F_{\text{fused}}$, contains spatially precise information from GroundingDINO, enriched with adaptively selected global feature information from CLIP.

\subsection{Modalities Enhancement with Modality Signature (MS)}
\begin{figure}[ht]
	\centering
	\includegraphics[width=0.95\linewidth]{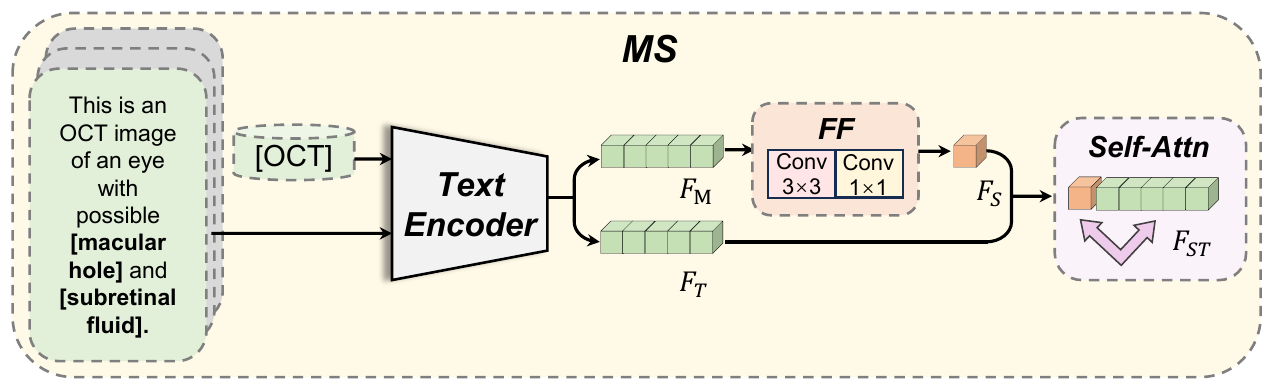}
	\caption{Overview of our proposed modalities fusion enhancement with modality signature (MS).} 
	\label{fig:ms}
\end{figure}

When fusing multiple imaging modalities, we face a unique challenge: modality ambiguity in textual descriptions. Specifically, many textual descriptions for ophthalmological diseases are generic, yet their visual manifestations differ drastically across various imaging modalities. For instance, the description "retinal thickening" signifies an increase in the thickness of retinal layers in an Optical Coherence Tomography (OCT) image. In contrast, within a color fundus photograph, this same condition might only appear as a vague, non-stereoscopic color anomaly, or it could even be imperceptible. If a model cannot perceive the specific imaging modality to which a textual description corresponds, it may erroneously apply textual priors from one modality to the visual features of another. This leads to feature confounding and subsequent performance degradation. 
Therefore, we design a Feature Fusion Module with a Modality Signature (MS) module in Fig.~\ref{fig:ms}. The core idea of this module is to first inject the corresponding imaging modality context into the text feature vector before its fusion with the image features. We achieve this by generating a compact yet informative Modality Signature, which is prepended to the original text feature vector. Concurrently, a self-attention mechanism is employed to enable the text features to attend to this modality signature. This enhanced text feature vector is then deeply fused with its corresponding image features, guiding the model toward more precise and context-aware multimodal learning.

% ----------------------------------------------------
% 1. Modality Signature Generation
% ----------------------------------------------------
\subsubsection*{Modality Signature Generation}
First, we encode the raw modality label into an initial feature vector ${F}_M \in \mathbb{R}^{D_{\text{label}}}$. 
To learn a more dense and expressive representation of the modality, we input ${F}_M$ into a small Feed-Forward Network (FFN), compressing it into a low-dimensional, dense vector. We term this output vector the modality signature ${F}_S \in \mathbb{R}^{D_{\text{sig}}}$:
\begin{equation}
    {F}_S = \text{FFN}({F}_M) = W_2 \cdot \text{ReLU}({W}_1 {F}_M + {b}_1) + {b}_2
    \label{eq:ffn_modality_signature}
\end{equation}
where ${W}_1, {b}_1, {W}_2, {b}_2$ are the learnable parameters of the FFN. The dimension of the signature, $D_{\text{sig}}$, is typically set to be small to efficiently convey modality information without introducing significant parametric overhead.

% ----------------------------------------------------
% 2. Signature Injection and Contextualization
% ----------------------------------------------------
\subsubsection*{Signature Injection and Contextualization}
After obtaining the modality signature ${F}_S$, we prepend it to the original text feature sequence ${F}_T$. The resulting combined sequence, ${F}_{ST}$, is then processed by a self-attention layer. This step allows every token in the original text to attend to the modality signature, thereby creating a holistically informed representation. This process yields the final modality-aware text feature, ${F}_{\text{ST'}}$:
\begin{align}
    {F}_{ST} &= \text{Concat}({F}_S, {F}_T) \label{eq:concat} \\
    {F}_{ST'} &= \text{SelfAttention}({F}_{ST}) \label{eq:self_attention_fusion}
\end{align}
The resulting feature, ${F}_{ST'}$, thus contains not only the semantic description of the disease but also the explicit context of its corresponding imaging modality.

\subsection{Integration of SAM Utilizing Bounding Box Prompts}
For the Segment Anything Model (SAM), we adopt an enhanced architecture based on MedSAM, tailored for ophthalmic multi-modal image segmentation tasks. MedSAM, a medical-adapted variant of SAM, incorporates domain-specific prior knowledge (e.g., anatomical structural constraints and pathological feature enhancement modules), significantly improving robustness in complex medical scenarios. In our framework, MedSAM serves as the core segmentation module, relying on spatial prompts provided by the preceding GroundingDINO stage. To enhance the accuracy and effectiveness of segmentation, we implement the following detailed processing steps: \textbf{Box Expansion}:  In the field of ophthalmology, images often exhibit small and intricate anatomical structures with low-contrast boundaries, making a robust and precise segmentation pipeline essential. To prevent the bounding boxes generated by Grounding DINO from being too tight, the box dimensions can be slightly expanded (by increasing the width and height by 5\%-10\% randomly) to ensure that the edges of the target object are fully covered, thereby improving segmentation quality.
\textbf{Multi-box Processing}: If Grounding DINO outputs multiple bounding boxes, each box can be sequentially fed into MedSAM to generate corresponding masks for multiple target objects. This approach ensures that all detected objects are accurately segmented, even in complex scenes with overlapping or closely positioned objects.
By incorporating these refinements, the combined use of Grounding DINO and MedSAM can achieve more precise and robust object detection and segmentation results.

\subsection{Loss Function}
Our loss function comprises three primary components: bounding box regression utilizing L1 and GIoU losses, as employed in GroundingDINO; segmentation loss incorporating Dice and cross-entropy losses, as utilized in MedSAM; and a cosine similarity loss between text embeddings and the cross-modal feature $\text{F}_\text{fused}$, leveraging the discriminative power of CLIP's contrastive loss.

\subsubsection*{Detection Losses}
The training    objective for the detection module is a combination of a classification loss $\mathcal{L}_{\text{cls}}$ and a bounding box regression loss $\mathcal{L}_{\text{bbox}}$:

\begin{equation}
\mathcal{L}_{\text{det}} = \lambda_{\text{cls}} \, \mathcal{L}_{\text{cls}} + \lambda_{\text{bbox}} \, \mathcal{L}_{\text{bbox}}
\end{equation}

where $\lambda_{\text{cls}}$ and $\lambda_{\text{bbox}}$ are hyperparameters that balance the contributions of the two terms.

\subsubsection*{Segmentation Losses}

To optimize the mask predictions, we employ a combination of the Dice loss $\mathcal{L}_{\text{dice}}$ and the binary cross-entropy (BCE) loss $\mathcal{L}_{\text{bce}}$:

\begin{equation}
\mathcal{L}_{\text{seg}} = \lambda_{\text{dice}} \, \mathcal{L}_{\text{dice}}(m_i, m_i^{\text{gt}}) + \lambda_{\text{bce}} \, \mathcal{L}_{\text{bce}}(m_i, m_i^{\text{gt}})
\end{equation}

where $m_i^{\text{gt}}$ denotes the ground truth mask for the $i$th object, and $\lambda_{\text{dice}}$ and $\lambda_{\text{bce}}$ are weighting parameters.

\subsubsection*{Contrastive Loss}
We employ a contrastive loss to enforce that the image and text embeddings corresponding to the same semantic content are close in the unified feature space, while those from mismatched pairs are far apart. The loss for a batch of $N$ $F_{fused}$ and text pairs is defined as:

\begin{equation}
\mathcal{L}_{\text{CLIP}} = -\frac{1}{N} \sum_{i=1}^N \log \frac{\exp\left( \text{sim}(\mathbf{F}_{fused}^{(i)}, \mathbf{F}_T^{(i)}) / \tau \right)}{\sum_{j=1}^{N} \exp\left( \text{sim}(\mathbf{F}_{fused}^{(i)}, \mathbf{F}_T^{(j)}) / \tau \right)}
\end{equation}

where  $\tau$ is temperature hyperparameter, the cosine similarity is defined as
\begin{equation}
\text{sim}(\mathbf{a}, \mathbf{b}) = \frac{\mathbf{a}^\top \mathbf{b}}{\|\mathbf{a}\|\|\mathbf{b}\|}
\end{equation}

\subsubsection{Joint Optimization and End-to-End Training}

To address modality imbalance, we assign a modality-specific weight $\omega_j$ to the loss computed for each modality $j$. Thus, given that each training batch contains only a single modality, for a set of modalities $\mathcal{M}$, the overall loss becomes:

\begin{equation}
\mathcal{L}^{\text{total}} = \sum_{j \in \mathcal{M}} \omega_j \, \mathcal{L}^{(j)}, \quad \text{with} \quad \omega_j \propto \frac{1}{\text{frequency}(j)}
\end{equation}

To jointly optimize the three modules (CLIP pretraining, detection, and segmentation) while accounting for modality imbalance, the overall training objective is defined as:

\begin{equation}
\mathcal{L}_{\text{total}} = \sum_{j \in \mathcal{M}} \omega_j \left( \mathcal{L}_{\text{CLIP}}^{(j)} + \mathcal{L}_{\text{det}}^{(j)} + \mathcal{L}_{\text{seg}}^{(j)} \right)
\end{equation}

\section{Experiments}
\newsavebox\CBox
\def\textBF#1{\sbox\CBox{#1}\resizebox{\wd\CBox}{\ht\CBox}{\textbf{#1}}}

\begin{figure*}[ht]
	\centering
    \includegraphics[width=.95\textwidth]{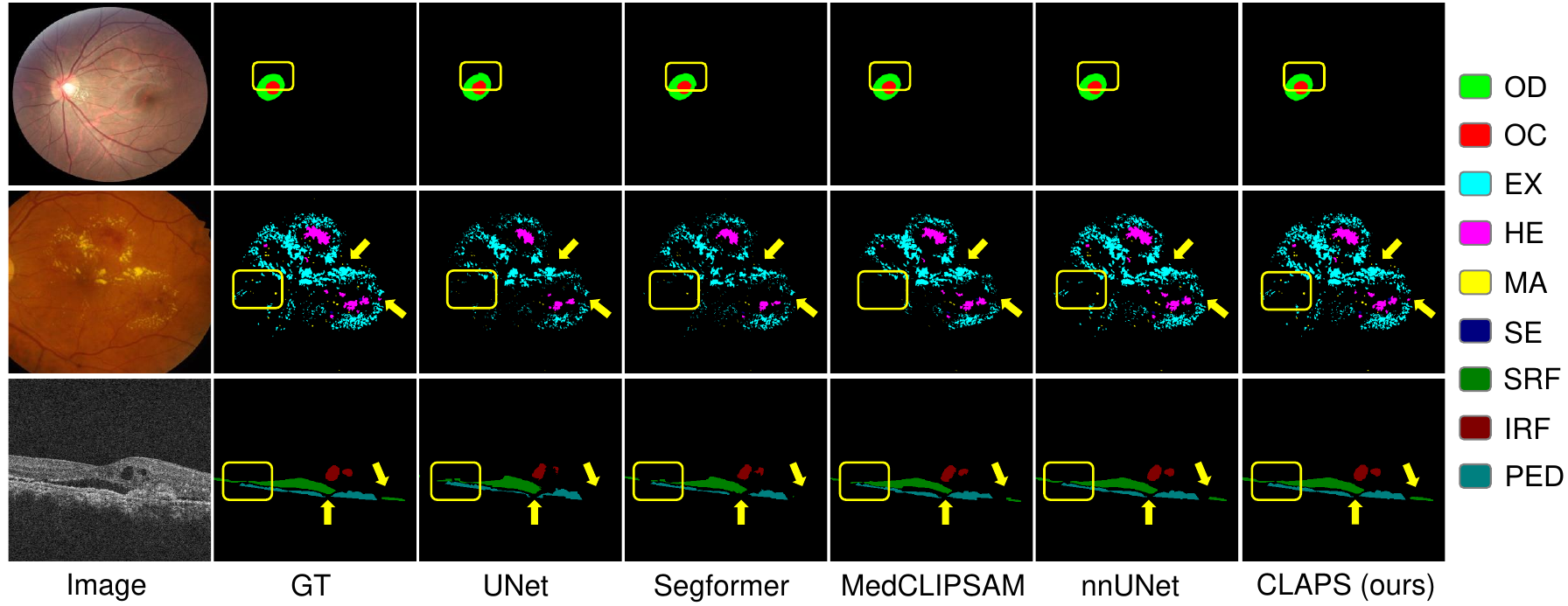}
    \caption{Visual comparative analysis of the efficacy of segmentation techniques on diverse data modalities}
    \label{fig:segresult}
\end{figure*}
\subsection{Setup Details}
\subsubsection{Dataset.}
To evaluate the model's capabilities, this study utilized fundus images and OCT B-scans. The fundus  datasets included optic cup and disc segmentation, specifically ORIGA \cite{zhang2010origa} and REFUGE \cite{orlando2020refuge}. Additionally, we employed datasets for segmenting  retinal lesions, such as hard exudates (EX), hemorrhages (HE), microaneurysms (MA), and soft exudates (SE). These lesion segmentation datasets consisted of E-optha \cite{decenciere2013teleophta}, IDRiD \cite{porwal2020idrid}, DDR \cite{li2019diagnostic}, and Retinal-lesion \cite{wei2021learn}. The OCT datasets focused on segmenting intraretinal fluid (IRF), subretinal fluid (SRF), and pigment epithelial detachment (PED) and comprised ReTouch (IRF, SRF, PED)\cite{bogunovic2019retouch}, AMD-SD (IRF, SRF, PED) \cite{hu2024amd}, OIMH (IRC, MH) \cite{ye2023oimhs}, AROI (IRF, PED)\cite{melinvsvcak2021aroi}, LUMHS (IRF) \cite{ahmed2022deep}, and IUMS (IRF) \cite{darooei2022dual}.

\subsubsection{Evaluation Metric.}
To illustrate the advantages of our approach, we conducted a comparative analysis of our models against UNet \cite{ronneberger2015u}, Segformer \cite{xie2021segformer}, nnUNet \cite{isensee2021nnu}, and MedCLIP-SAM \cite{koleilat2024medclip}. The evaluation metric employed was the Dice Metric.

\subsubsection{Experimental Settings.}
The  MedCLIP model was pre-trained using PyTorch on 2 Nvidia RTX 4090 24GB GPUs, with a total training duration of 200 epochs, including a 50-epoch warm-up period, and a learning rate set at 5e-5. For bounding box detection, all images are first padded into squares based on their longest side and then resized to 224x224. In the MedSAM segmentation phase, the images are kept at their original size.

\subsection{Evaluation and Results}

We standardized and combined the training sets of all datasets into a unified format for training. After adequate training, we evaluated the test sets of each dataset individually. Fig.~\ref{fig:segresult} visually compares segmentation results across methods. UNet and Segformer perform well in segmenting large areas like the optic disc and cup but struggle with small targets such as microaneurysms. nnUNet shows superior performance but still misses some small targets due to information loss from downsampling. MedCLIP-SAM \cite{koleilat2024medclip} demonstrates superior performance at a macroscopic scale, yet its efficacy diminishes significantly at a microscopic level. The primary reason for this limitation is the inability of the CAM approach to precisely localize minute pathological lesions. In contrast, our model achieves fully automatic segmentation, combining the strengths of different models and excelling in small region segmentation. This is attributed to the CLIP model's extensive data learning, enabling it to effectively capture disease-specific areas in retinal images.

\begin{table}[h]
\centering
\caption{Quantitative Analysis Results (Dice) on CFP datasets: ORIGA, REFUGE, DIRiD}
\label{tab:fundus_results_1}
\resizebox{.98\linewidth}{!}{
\begin{tblr}{
  cell{1}{2} = {c=2}{c},
  cell{1}{4} = {c=2}{c},
  cell{1}{6} = {c=4}{c},
  % vline{2-3,5,7,9,13} = {1}{},
  vline{2,4,6} = {1-7}{},
  hline{1,3,7-8} = {-}{},
  hline{2} = {2-10}{},
}
Dataset                                  & ORIGA  &        & REFUGE   &        & IDRiD  &        &        &        \\
                                         & OD     & OC     & OD       & OC     & EX     & HE     & SE     & MA     \\
UNet \cite{ronneberger2015u}             & 95.37  & 88.93  & 93.17    & 88.42  & 73.18  & 61.92  & 64.01  & 32.95  \\
Segformer \cite{xie2021segformer}        & 95.98  & 89.81  & 95.07    & 89.52  & 76.25  & 64.66  & 69.58  & 40.83  \\
nnUNet \cite{isensee2021nnu}             & 96.68  & 95.09  & \textBF{97.51} & 95.66  & 80.56  & \textBF{67.51} & 73.04  & 46.12  \\
MedCLIP-SAM \cite{koleilat2024medclip}   & 96.19  & 92.51  & 96.99    & 93.42  & 78.91  & 66.13  & 70.42  & 44.46  \\
CLAPS (ours)                             & \textBF{96.78} & \textBF{95.21} & 97.18    & \textBF{95.81} & \textBF{80.72} & 66.79  & \textBF{73.07} & \textBF{46.50} 

\end{tblr}
}
\end{table}

\begin{table}[h]
\centering
\caption{Quantitative Analysis Results (Dice) on CFP Datasets: Eopha, DDR, Retina-lesion }
\label{tab:fundus_results_2}
\resizebox{.98\linewidth}{!}{
\begin{tblr}{
  cell{1}{2} = {c=1}{c},
  cell{1}{3} = {c=4}{c},
  cell{1}{7} = {c=4}{c},
  % vline{2-3,5,7,9,13} = {1}{},
  vline{2,3,7} = {1-7}{},
  hline{1,3,7-8} = {-}{},
  hline{2} = {2-11}{},
}
Dataset                                   & EOptha     & DDR    &        &        &        & Retina-Lesion  &        &        &       \\
                                          & EX         & EX     & HE     & SE     & MA     & EX             & HE     & SE     & MA    \\
UNet \cite{ronneberger2015u}              & 79.49      & 53.51  & 42.88  & 40.58  & 15.72  & 72.92          & 58.34  & 52.83  & 32.78 \\
Segformer \cite{xie2021segformer}         & 84.01      & 57.63  & 46.59  & 32.22  & 24.85  & 73.82          & 60.94  & 57.63  & 36.97 \\
nnUNet \cite{isensee2021nnu}              & 91.38      & 60.52  & 47.74  & 49.86  & 27.82  & 74.45          & 61.53  & \textBF{57.83} & 39.48 \\
MedCLIP-SAM \cite{koleilat2024medclip}    & 88.50      & 58.05  & 51.52  & 50.94  & 26.14  & 73.15          & 59.93  & 56.61  & 38.48 \\
CLAPS (ours)                              & \textBF{92.04}     & \textBF{61.26} & \textBF{52.26} & \textBF{51.27} & \textBF{28.25} & \textBF{75.26}         & \textBF{62.25} & 56.92  & \textBF{40.12}

\end{tblr}
}
\end{table}

\begin{table}[h]
  \centering
  \caption{Quantitative Analysis Results (Dice) on OCT Datasets: ReTouch, AMD-SD}
  \label{tab:results_oct_1}
  \resizebox{.98\linewidth}{!}{
  \begin{tblr}{
    cell{1}{2} = {c=3}{c},
    cell{1}{5} = {c=3}{c},
    % vline{2-3,6,9,11,13} = {1}{},
    % vline{2,5,8,10,12-13} = {1}{},
    vline{2,5} = {1-7}{},
    hline{1,3,7-8} = {-}{},
    hline{2} = {2-8}{},
  }
  Dataset                                  & ReTouch  &        &        & AMD-SD  &        &        \\
                                           & IRF      & SRF    & PED    & IRF     & SRF    & PED    \\
    UNet \cite{ronneberger2015u}           & 61.03    & 70.05  & 73.04  & 78.72   & 78.71  & 79.66  \\
    Segformer \cite{xie2021segformer}      & 69.06    & 74.08  & 77.03  & 81.74   & 79.02  & 81.93  \\
    nnUNet \cite{isensee2021nnu}           & 81.07    & \textBF{83.05} & 85.06  & 83.84   & 82.86  & 84.24  \\
    MedCLIP-SAM \cite{koleilat2024medclip} & 78.04    & 82.06  & 84.03  & 82.30   & 82.14  & 83.83  \\
    CLAPS (ours)                           & \textBF{81.20}   & 82.17  & \textBF{86.19} & \textBF{85.18}  & \textBF{84.18} & \textBF{85.77}

  \end{tblr}
  }
  \end{table}

\begin{table}[h]
  \centering
  \caption{Quantitative Analysis Results (Dice) on OCT Datasets: OIMH, AROI, LUMHS}
  \label{tab:results_oct_2}
  \resizebox{.98\linewidth}{!}{
  \begin{tblr}{
    cell{1}{2} = {c=2}{c},
    cell{1}{4} = {c=2}{c},
    % vline{2-3,6,9,11,13} = {1}{},
    % vline{2,5,8,10,12-13} = {1}{},
    vline{2,4,6,7} = {1-7}{},
    hline{1,3,7-8} = {-}{},
    hline{2} = {2-8}{},
  }
  Dataset                                   & OIMH   &        & AROI   &        & LUMHS  & IUMS  \\
                                            & IRC    & MH     & IRF    & PED    & IRF    & IRF   \\
    UNet \cite{ronneberger2015u}            & 86.55  & 87.15  & 86.35  & 87.95  & 84.33  & 74.68 \\
    Segformer \cite{xie2021segformer}       & 87.08  & 88.72  & 87.02  & 89.25  & 86.94  & 76.78 \\
    nnUNet \cite{isensee2021nnu}            & 90.25  & \textBF{91.35} & 88.45  & \textBF{90.96} & 88.71  & 80.29 \\
    MedCLIP-SAM \cite{koleilat2024medclip}  & 90.66  & 89.75  & 88.85  & 89.55  & 88.37  & 79.97 \\
    CLAPS (ours)                            & \textBF{93.21} & 91.17  & \textBF{89.18} & 90.19  & \textBF{90.20} & \textBF{82.21}

  \end{tblr}
  }
  \end{table}
To comprehensively evaluate the performance of our proposed CLAPS, we conducted extensive quantitative assessments on a diverse collection of 12 public datasets, spanning two key ophthalmic imaging modalities: color fundus photography (CFP) and optical coherence tomography (OCT). The results benchmark CLAPS against several established segmentation models, including the highly competitive nnUNet, which was trained as a specialist model on each individual dataset.
In the domain of fundus image segmentation (Tables ~\ref{tab:fundus_results_1} and ~\ref{tab:fundus_results_2}), our CLAPS model demonstrates consistently superior or highly competitive performance. For larger anatomical structures, such as the Optic Disc (OD) and Optic Cup (OC) in the ORIGA and REFUGE datasets, CLAPS achieves SOTA results, performing on par with or marginally outperforming the specialist nnUNet model (e.g., 96.78 vs. 96.68 for OD in ORIGA). The advantage of our unified model becomes more pronounced when segmenting smaller, more challenging pathological lesions. Across the IDRiD, DDR, and Retina-Lesion datasets, CLAPS consistently sets a new state-of-the-art for segmenting Exudates (EX), Hemorrhages (HE), and Microaneurysms (MA). This pattern suggests that our model's unified training strategy enables it to learn robust, generalizable features that are particularly effective for localizing fine-grained, clinically relevant biomarkers.

This trend of robust performance extends to the OCT modality, as shown in Tables ~\ref{tab:results_oct_1} and ~\ref{tab:results_oct_2}, further validating the versatility of CLAPS. Our model achieves the highest Dice scores in 9 out of the 12 evaluated OCT segmentation tasks. On the ReTouch and AMD-SD datasets, CLAPS demonstrates superior segmentation of key fluid biomarkers, achieving top scores for IRF, SRF, and PED in five out of six categories. This leadership continues across the OIMH, AROI, LUMHS, and IUMS datasets. Notably, CLAPS shows a substantial improvement in segmenting IRC in the OIMH dataset (93.21 vs. nnUNet's 90.25) and IRF in the LUMHS (90.20 vs. 88.71) and IUMS (82.21 vs. 80.29) datasets. While nnUNet maintains a slight edge in a few specific categories, our model’s consistently high scores across a greater number of tasks highlight its strong overall capability.
In summary, the comprehensive quantitative results demonstrate that our unified model, CLAPS, not only competes with but also often surpasses specialized, single-dataset models. Its exceptional performance, particularly in segmenting fine-grained pathological features in fundus images and various fluid regions in OCT scans, validates the effectiveness of our approach. 

\begin{table}[h]
\centering
\caption{Evaluating the Impact of Individual Modules on Fundus Datasets.}
\label{tab:ablation_1}
\resizebox{.98\linewidth}{!}{
\begin{tblr}{
  colspec = {Q[c]Q[c]Q[c]Q[c]Q[c]Q[c]Q[c]},
  vline{2} = {1-7}{},
  hline{1,2,7} = {-}{},
}
Dataset                       & Refuge & ORIGA & EOphtha & IDRiD & DDR   & Retina-Lesion \\
+P                            & 90.56  & 92.30 & 75.50   & 66.94 & 48.30 & 65.88         \\
+GD                           & 93.44  & 95.29 & 85.93   & 72.90 & 54.25 & 70.00         \\
+GD+CLIP                      & 94.81  & 95.52 & 88.45   & 73.96 & 54.86 & 70.49         \\
MedSAM(PreTr)+GD+CLIP         & 96.02  & 95.91 & 90.81   & 75.65 & 56.02 & 71.96         \\
MedSAM(Train)+GD+CLIP         & 96.91  & 96.14 & 92.05   & 77.38 & 57.94 & 73.02          
\end{tblr}
}
\end{table}

\begin{table}[h]
\centering
\caption{Evaluating the Impact of Individual Modules on OCT Datasets.}
\label{tab:ablation_2}
\resizebox{.98\linewidth}{!}{
\begin{tblr}{
  colspec = {Q[c]Q[c]Q[c]Q[c]Q[c]Q[c]Q[c]},
  vline{2} = {1-7}{},
  hline{1,2,7} = {-}{},
}
Dataset                        & ReTouch & AMD-SD & OIMH  & AROI  & LUMHS & IUMS  \\
+P                             & 69.99   & 78.89  & 86.90 & 87.37 & 82.25 & 71.74 \\
+GD                            & 79.83   & 82.74  & 89.59 & 88.53 & 87.88 & 76.95 \\
+GD+CLIP                       & 82.01   & 83.11  & 90.24 & 89.03 & 88.34 & 79.70 \\
MedSAM(PreTr)+GD+CLIP          & 82.83   & 85.00  & 92.03 & 89.33 & 90.07 & 82.17 \\
MedSAM(Train)+GD+CLIP          & 84.40   & 85.07  & 92.46 & 89.99 & 90.24 & 82.26  
\end{tblr}
}
\end{table}

\subsection{Ablation Study}
%%%% 
In this section, we conducted ablation experiments to evaluate the effectiveness of different modules. Our baseline model was MedSAM, pre-trained on a general medical dataset. The additional modules we incorporated include: (1) integrating text prompts (P) into MedSAM and fine-tuning it on retinal segmentation datasets; (2) combining GroundingDino (GD) with MedSAM, utilizing the bounding boxes obtained from GD as prompts for segmentation; and (3) further enhancing the model by integrating CLIP image and text encoders pre-trained on a large-scale retinal dataset. Finally, we compared the segmentation performance of MedSAM fine-tuned on a retinal lesion segmentation dataset with that of MedSAM trained end-to-end within the complete model. The results in Tables ~\ref{tab:ablation_1} ~\ref{tab:ablation_2} clearly demonstrate that simply adding text prompts to MedSAM and fine-tuning it yields significantly inferior performance compared to the combined MedSAM-GD model without fine-tuning. Additionally, the CLIP module substantially improved segmentation performance, particularly on datasets with numerous lesions and imbalanced lesion distributions, such as IDRiD and ReTouch. Moreover, end-to-end training of MedSAM further enhanced the model's performance on imbalanced datasets.

%%%%%%%%%%%%%%%%%%%%%%%%%%%%%%%%%%%%%%%%%%%%%%%%%%%%%%

%%%%%%%%%%%%%%%%%%%%%%%% Conclusion %%%%%%%%%%%%%%%%%%

%%%%%%%%%%%%%%%%%%%%%%%%%%%%%%%%%%%%%%%%%%%%%%%%%%%%%%

\section{Conclusions}

CLAPS introduces a unified and automated approach to multi-modal retinal image segmentation, addressing key limitations of existing methods. By leveraging GroundingDINO for automatic ROI detection and SAM for prompt-guided segmentation, CLAPS eliminates the need for manual intervention while ensuring precise lesion delineation. Additionally, CLIP-based pretraining enables the model to project diverse retinal images into a shared feature space, mitigating data imbalance and enhancing generalizability. Extensive experiments across multiple datasets demonstrate CLAPS’s superior performance and adaptability, making it a promising foundation model for retinal imaging. This work paves the way for more robust and scalable segmentation frameworks in ophthalmic analysis.

\bibliographystyle{IEEEtran}
\bibliography{IEEEabrv,bibliography}

\end{document}